# Wireless Sensor Networks anomaly detection using Machine Learning: A Survey

Ahsnaul Haque[1], Md Naseef-Ur-Rahman Chowdhury[1], Hamdy Soliman[1], Mohammad Sahinur Hossen[1], Tanjim Fatima[1], and Imtiaz Ahmed[1]

New Mexico Tech, 801 Leroy PL, Socorro, NM, USA,
ahshanul.haque@student.nmt.edu, naseef.chowdhury@student.nmt.edu,
hamdy.soliman@nmt.edu, mohammad.hossen@student.nmt.edu,
tanjim.fatima@student.nmt.edu, imtiaz.ahmed@student.nmt.edu

Abstract. Wireless Sensor Networks (WSNs) have become increasingly valuable in various civil/military applications like industrial process control, civil engineering applications such as buildings' structural strength monitoring, environmental monitoring, border intrusion, IoT (Internet of Things), and healthcare. However, the sensed data generated by WSNs is often noisy and unreliable, making it a challenge to detect and diagnose anomalies. Machine learning (ML) techniques have been widely used to address this problem by detecting and identifying unusual patterns in the sensed data. This survey paper provides an overview of the state of-the-art applications of ML techniques for data anomaly detection in WSN domains. We first introduce the characteristics of WSNs and the challenges of anomaly detection in WSNs. Then, we review various ML techniques such as supervised, unsupervised, and semi-supervised learn ing that have been applied to WSN data anomaly detection. We also compare different ML-based approaches and their performance evalu ation metrics. Finally, we discuss open research challenges and future directions for applying ML techniques in WSNs sensed data anomaly detection.

Keywords: Wireless Sensor Network, Anomaly Detection, Machine Learn ing[16], Survey, Energy Efficiency, Hybrid Networks, Techniques, Algo rithms, Data, Performance Metrics

## 1 Introduction

Wireless Sensor Networks (WSNs) are widely used in various types of applications, including environmental monitoring, Internet-Of-Things (IoT), healthcare, security[18], and industrial control. Some WSNs[20]
consist of numerous tiny sensors that are distributed in an area to col lect data and send it to a base station. However, due to the resource constrained nature of sensor nodes, WSNs[21] face several challenges, such as limited energy, computational power, and memory, making it difficult to make decisions and analyze data locally[1, 2, 31, 33].

Anomaly detection is a critical task in WSNs since it helps detect unusual events and abnormal behavior in the sensed data. Anomalies may indicate a malfunction in the system's sensors, equipment fail ure, or potential security threats, which need to be addressed imme diately, and appropriately. However, traditional rule-based anomaly detection techniques may not be suitable for WSNs since they re quire predefined rules that are challenging to design for complex WSNs systems with high-dimensional data[3, 5].

ML has emerged as a promising technique for anomaly detection in WSNs. ML algorithms can learn from the sensor data and discover patterns that are indicative of normal and abnormal behavior, with out the need for manual rule definition. Moreover, ML techniques can adapt to changes in the system and the environment and pro vide accurate and timely detection of anomalies, thus improving the efficiency and reliability of the WSNs[4]. A common ML-based tech nique is to use the Ensemble Learning method[11, 12], where multiple ML models are used to train and test, then pick the best-performing models. This approach is not only effective in WSN but also widely used in many other anomalies/malware detection[11][12].

Given the growing interest in using ML for anomaly detection in WSNs, it is essential, for life-saving critical applications, to sur vey the state-of-the-art techniques and identify the challenges and opportunities in this field. This paper aims to provide a comprehen sive survey of ML-based anomaly detection methods in WSNs. The survey covers various ML algorithms, including supervised[3], unsu pervised[13], semi-supervised[5], and deep learning[7], and discusses their advantages in modeling sensed data anomalies and limitations in WSNs. The paper also discusses the challenges of deploying ML based anomaly detection systems in WSNs, such as limited resources, collected data heterogeneity, and privacy/security concerns, and sug gests potential solutions to address these challenges. Overall, this survey can serve as a useful reference for researchers and practition ers working on anomaly detection in WSNs using ML techniques[6- 9,20-24,27,28,30-32,34-38,40,41].

The rest of this paper is organized as follows. In the second sec tion, we will discuss the most recent Criterion of Anomalies. The fol lowing section will discuss the Anomaly Detection Techniques used to date. In Section 4, we will compare the different anomaly detection approaches for WSNs. In the final section, we conclude by shedding some light on the most useful ML models to be deployed in the field of anomaly detection of collected WSNs sensed data.

## 2 Classification Criterion of Anomalies

In recent years, WSNs have emerged as a captivating research field due to their ability to monitor vast regions, reach remote and haz ardous locations, respond in real time, and their relative simplicity of use[34, 36]. This technology has opened up a whole new world of possibilities for scientists. In addition to the aforementioned impor tant civil applications, WSNs have already been utilized in many other various military activities such as surveillance, target recogni tion, and environmental. These sensor networks typically consist of numerous small, inexpensive nodes distributed across a wide area, as shown in Figure 1.

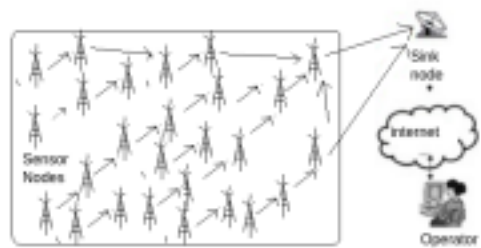

Fig. 1. An example architecture of WSNs[2]

WSNs not only have the ability to sense, compute and coordi nate their activities but also to communicate their results to end users, making them revolutionary in data collection across various domains[23]. However, the unique research and engineering chal lenges that arise during the deployment and design of these networks must be considered, as well as the limitations of their software devel opment[37, 42]. These limitations include their large intended area of deployment range, communication obstacles, random and hazardous deployment, high components' failure rate, and limited computa tional and energy/battery power. To ensure better critical decision-making, it is crucial to maintain the quality of collected sensor data. Although cryptographic and key management techniques are used to protect the security of sensor nodes from intruder attacks, they are not sufficient to ensure the reliability and integrity of their sensed data[35, 41].

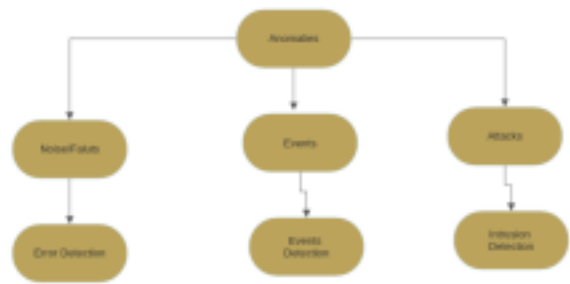

Fig. 2. Identification of Anomalies in WSN [22]

Thus, outlier detection techniques have been developed to iden tify any abnormal behavior in sensor data streams. WSNs are par ticularly susceptible to outliers due to several factors. Such factors include their use of weak and vulnerable sensors to collect data in real-world applications and their battery-powered nature. The po tential accumulation of errors when numerous sensors are used over wireless media, and the vulnerability of unguarded sensors in criti cal security and military applications to manipulation by intruders. Therefore, outlier detection is an integral part of any perilous data processing task that utilizes

WSNs. The following subsections out line the fundamental concepts, sources, and requirements of outlier detection in WSNs[38].

In WSNs, anomalies refer to any unusual, abnormal, or unex pected behavior or events in the collected sensor data stream that de viate from the expected or normal patterns. Anomalies can be caused by various factors, such as faulty sensors, environmental changes, malicious attacks, or random fluctuations of the sensed data, due to external random/asynchronous events in the terrain.

Anomalies can be categorized into three major types, namely: noise, event, and attack anomaly sources. We pictured the classifi cation of anomalies in Figure 3.

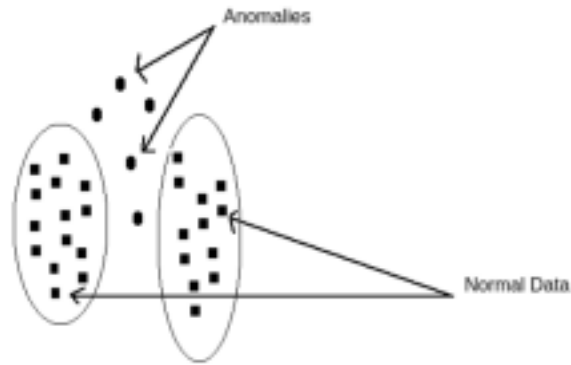

Fig. 3. An example Anomalies with 2-dimensional dataset[5]

Noise or errors anomaly in WSNs refer to measurement inaccu racies or data sensed from sources like faulty or malfunctioning sen sors[28]. Outliers resulting from errors can occur frequently and are typically represented by a data point that differs significantly from the rest of the collected dataset. They can arise due to various envi ronmental factors, including bad deployment due to difficulties and harsh conditions [29]. To ensure data quality, both detected noisy and erroneous data should be eliminated or corrected, if possible[39].

Event anomaly sources in WSNs are defined as sudden changes in the real-world state, such as fire detection [31],

earthquakes, weather changes, and air pollution [30]. Outliers caused by such anomaly sources tend to have a significantly lower probability of occurrence compared to those caused by errors. Such outliers typically last for a relatively long period of time and can alter the historical pattern of sensor data[24]. Removing event outliers from the dataset can lead to the loss of crucial information about the events [32]. Outliers that are similar in size to random errors can only be identified through the application of outlier tests.

In WSNs, malicious attack anomaly sources are associated with network security, and researchers in [33] have addressed this issue. Due to the unattended nature of the deployment of the sensors, intruders can gain access to and control, damage, and/or hijack spe cific nodes to launch attacks, which can deplete the network's limited resources or inject false and corrupted data. Malicious attacks can be broadly classified into two categories: passive and active attacks. Passive attacks involve obtaining data interchanged in the network without interrupting communication, while active attacks aim to dis rupt the normal functioning of the network[40].

## 3 Anomaly Detection Techniques in WSN

### 3.1 Classification of the Approaches

There are three main types of modeling applications categories of ML[14] for anomaly detection in WSNs namely: supervised, unsu pervised, and semi-supervised. The appropriate category should be chosen based on the WSN anomaly detection task's specific require ments, as each category has its own advantages and disadvantages.

Based on the characteristics of the Data-set and the user's needs we can use different ML approaches. Figure 4 is showing the demon stration of the ML approaches in detecting the anomalies for WSNs.

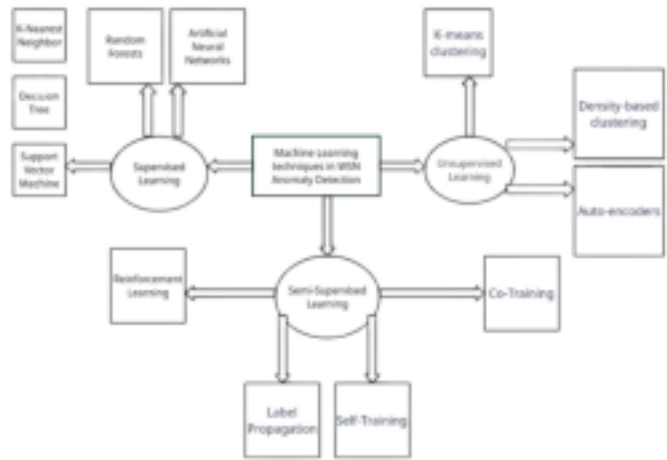

Fig. 4. ML Approaches to detect anomalies in WSN[22]

SVM, KNN, Random Forest, Decision Tree, ANN, K-means Clus tering, Density-Based Clustering, Auto-Encoders, Reinforcement Learn ing, Self Training, Co-Training, and Label Propagation, are popular ML algorithms used for anomaly detection in WSN[1, 4, 5, 7]. Each algorithm has its own strengths and weaknesses depending on the specific use case. Next, we will shed some light on each of such ML algorithms to justify its utilization in some applications.

WSN anomaly detection using Machine Learning: A Survey 7

In the next three sub-sections, we will survey the detailed liter ature works in each of the aforementioned training/learning cate gories: supervised, unsupervised, and semi-supervised.

3.2 Supervised Learning Approaches

Labeled training data are used by supervised learning[3, 5, 7–9] al gorithms, and along with teacher supervision, to learn the normal network behavior for later identification of anomalies based on de viations from the normal learned behavior. When labeled data for training are available and the network's normal behavior class is well-defined, this strategy is appropriate. Next, we will briefly intro duce some of the prominent supervised ML models utilized in WSNs data classification for anomaly detection and explore some related literature works.

3.2.1 A brief Introduction for Supervised ML models

SVM[15] is a statistical-based[17] approach that is effective in han dling high-dimensional data and can separate classes with large mar gins, but it requires tuning of hyperparameters and can be compu tationally expensive[3, 8].

KNN[22] is a clustering-based method that is useful for detecting local anomalies and is simple to implement, but it is sensitive to the choice of distance metric and can be computationally expensive.

Random Forests[9, 10] is a ML-based approach that is useful for handling noisy data, can identify feature importance, and is paral lelizable, but it has limited interpretability and can overfit the data.

ANN[9] is another ML-based approach that is good at handling non-linear relationships, can handle complex data, and is paralleliz able, but requires careful selection of architecture and hyperparam eters and can be computationally expensive.

Decision Tree[22] is a ML-based approach that is useful for handling noisy data, is easy to interpret, and can handle missing values, but it is prone to overfitting and sensitive to the choice of split criteria. Ultimately, the choice of algorithm depends on the specific characteristics of the data and the requirements of the application.

3.2.2 Literature work in Supervised Learning

In [1], the authors reviewed the challenges faced by WSNs and pro posed the use of ML techniques to improve their energy efficiency and detect anomalies in their collected data. The authors provide a comprehensive literature review of ML algorithms that have been ap plied to WSNs, including K-Means, ANN, Decision Trees, SVM[19], and Bayesian Networks[22, 2, 5, 13]. The author presents two case studies that demonstrate the effectiveness of ML techniques for en ergy efficiency and anomaly detection in WSNs. Overall, the paper contributes to the growing interest of research on the use of ML for WSNs and highlights the potential of these techniques for improving the performance and reliability of WSNs.

The authors in [4] proposed a method for detecting spatial anoma lies in sensor networks using neighborhood information. The authors provide a literature review of related work in anomaly detection in WSNs and highlight the challenges of detecting spatial anomalies due to the complexity of spatial data. The paper presents the proposed method, which uses neighborhood information to identify anomalies in spatial data (consisting of latitude and longitude). The authors demonstrate the effectiveness of their method through experiments conducted on simulated and real-world data sets. Overall, the pa per contributes to the growing interest in research on using sensor networks for anomaly detection and highlights the potential of using neighborhood information to improve such systems' accuracy and reliability.

The authors in [5] proposed a new approach to anomaly detec tion in WSNs using support vector data description (SVDD). The authors provide a literature review of related work in the field of anomaly detection in WSNs and highlight the limitations of existing approaches, such as high false alarm rates and low detection accu racy. The paper presents the proposed method, which uses SVDD to construct a boundary around normal data and detect anomalies outside this boundary. The authors demonstrate the effectiveness of their method through experiments conducted on real-world data sets, showing that their method outperforms existing methods in terms of detection accuracy and false alarm rate. Overall, the paper contributes to the field of smart anomaly detection in WSNs and highlights the potential of using SVDD to improve such systems' accuracy and reliability.

The author in [7] proposed a data-driven approach for hyperpa rameter optimization of one-class SVMs for anomaly detection in WSNs. The authors provided a literature review of related work on anomaly detection in WSNs and highlight the importance of hyper parameter optimization for improving the accuracy and efficiency of SVM-based methods. The paper presents the proposed method, which uses a genetic algorithm to optimize the hyperparameters of the one-class SVM model. The authors demonstrate the effective ness of their method through experiments conducted on real-world data sets, showing that their method outperforms existing methods in terms of detection accuracy and computational efficiency. Over all, the paper contributes to smart anomaly detection in WSNs and highlights the potential of using data-driven approaches for hyper parameter optimization of SVM-based methods.

The authors in [3] proposed a new algorithm for anomaly detec tion in WSNs using a combination of density-based spatial cluster ing of applications with noise (DBSCAN) and SVM. The authors provided a literature review of related work on anomaly detection in WSNs and highlight the challenges faced in implementing such systems, including limited computational resources, communication bandwidth, and energy constraints. The paper presents the proposed DBSCAN algorithm and demonstrates its effectiveness in detecting anomalies in WSNs using simulations. Overall, the paper contributes to the smart use of ML techniques for anomaly detection in WSNs and highlights the potential of combining DBSCAN and SVM for improving the accuracy and reliability of such systems.

The authors in [8] presented a comprehensive literature review of existing intrusion detection systems (IDSs) for WSNs and investi gated the applicability of computational intelligence techniques for enhancing the performance of IDSs. The authors discussed the chal lenges and requirements of intrusion detection in WSNs, including limited resources, wireless communication, and distributed deploy ment. Then, they reviewed the use of various soft-computational intelligence techniques for IDSs, such as ANNs, decision trees, fuzzy logic, and genetic algorithms. The paper also provides an overview of existing benchmark datasets and evaluation metrics for IDSs in WSNs. Overall, the paper highlights the potential of using soft computing intelligence techniques for improving the accuracy.

The authors in [9] proposed a distributed anomaly detection scheme based on autoencoder neural networks for WSNs. The authors aim to address the limitations of existing centralized anomaly detection methods that may not be suitable for large-scale WSNs due to band width constraints, power consumption, and privacy concerns. The proposed scheme involves training autoencoder neural networks at each sensor node to learn the normal behavior of the local envi ronment and detect anomalies based on reconstruction errors. The paper also discusses the implementation of the proposed scheme on a testbed and evaluates its performance using various metrics. The results show that the proposed scheme achieves high accuracy in detecting anomalies while maintaining low communication overhead and energy consumption. Hence, such an approach will be suitable for large-scale WSNs in IoT applications. Overall, the paper provides a valuable contribution to the field of distributed anomaly detection in WSNs and demonstrates the potential of using autoencoder neural networks for improving the efficiency and effectiveness of anomaly detection in IoT applications. The authors in [10] proposed a deep learning-based approach for developing middleware for WSNs. The authors aim to address the limitations of existing middleware ap proaches that may not be able to handle the complexity and het erogeneity of WSNs, leading to low accuracy and reliability. The proposed approach involves training a deep neural network (DNN) to predict the behavior of the WSN based on historical data and using the predictions to adjust the middleware parameters in real time. The paper also discusses the implementation of the proposed approach on a testbed and evaluates its performance using various metrics. The results show that the proposed approach achieves high accuracy and reliability in adapting the middleware to the dynamic behavior of the WSN, making it suitable for various WSN appli cations. Overall, the paper provides a valuable contribution to the field of WSN middleware development and demonstrates the poten tial of using deep learning for improving the accuracy and reliability of WSN middleware.

3.3 Unsupervised Learning Approaches

The assumption that anomalies are data points that are significantly distinct from most of the data points serves as the foundation for unsupervised learning algorithms[6, 25–27]. Hence, no need for a su pervisor teacher, the ML model will cluster anomaly data separately from normal data, even with unlabeled data based on the data qual ity. When labeled data is unavailable or the network's normal be havior is unclear, this method is appropriate. Next, we will briefly introduce some of the prominent unsupervised ML models utilized in WSNs data classification for anomaly detection and explore some related literature works.

3.3.1 Brief Introduction to unsupervised ML models

K-means clustering[6] is a popular unsupervised clustering method that is used for identifying clusters of data points with similar charac teristics[6, 26, 27]. It is useful for detecting global anomalies and can handle large datasets, but it requires the selection of the appropri ate number of clusters and can be sensitive to outliers. Density-based clustering, on the other hand, is useful for detecting local anomalies and does not require specifying the number of clusters beforehand[2, 25]. However, it can be computationally expensive and sensitive to the choice of parameters.

Auto-encoders[26] are a type of neural network that is used for unsupervised feature learning and anomaly detection. They are use ful for detecting non-linear relationships in the data and can handle complex data, but they require a careful selection of architecture and hyper-parameters and can be computationally expensive. Ultimately, the choice of technique depends on the specific characteristics of the data and the requirements of the application[25, 26].

Density-based clustering[26] is a data clustering technique that identifies clusters of points based on the density of their distri bution. It is particularly useful for identifying clusters of arbitrary shape and can handle noise and outliers effectively.

3.3.2 Literature work in Unsupervised Learning

In [2], the authors presented a method for detecting anomalies in a WSN used in a smart home system. The authors propose using sta tistical analysis and ML techniques to detect abnormal behavior in

12 A. Haque, N. Chowdhury, H. Soliman et al.

the network, which can be used to identify security breaches, faults, and other anomalies. The paper provides a literature review of re lated work in the field of anomaly detection in WSNs and highlights the challenges of implementing such systems in smart home environ ments. The authors present the results of experiments conducted on a prototype smart home system, which demonstrate the effectiveness of their proposed method for detecting anomalies in WSNs. Overall, the paper contributes to the growing interest in the use of WSNs in smart home environments and highlights the potential of ML tech niques for improving the security and reliability of such systems.

The authors in [6] proposed an attention-based multi-filter long short-term memory (AMF-LSTM)-based deep learning strategy for network anomaly detection. The authors present a review of related research on anomaly detection in networks and draw attention to the drawbacks of existing approaches, such as the need for domain knowledge and the inability to identify anomalies that were not pre viously known, hence the use of unsupervised learning. The proposed approach, which makes use of AMF-LSTM to capture the temporal and spatial correlations that exist in network traffic, is presented in the paper, along with a focus on the essential characteristics for anomaly detection. Experiments on real-world data sets used by the authors to demonstrate the efficacy of their method show that, in terms of detection accuracy and false alarm rate, it outperforms the peers. In general, the paper highlights the potential of using AMF LSTM to improve the accuracy and dependability of such systems and adds to the growing interest in research on the use of deep learn ing techniques for anomaly detection in networks.

The authors in [25] proposed a new approach for detecting outliers in WSNs using k-means clustering and lightweight methods. They argue that outlier detection is essential for improving network perfor mance and reliability, and that existing methods have limitations in terms of computational complexity and memory requirements. The authors introduce a two-phase approach that divides sensor nodes into clusters based on similarity and then applies lightweight out lier detection methods to each cluster. To evaluate the effectiveness of the proposed approach, the authors conducted experiments using real-world data from a WSN. The results showed that the proposed approach outperformed existing outlier detection methods in terms of both detection accuracy and computational efficiency.

The authors in [26] presented two novel approaches for detecting outliers in WSNs using density-based methods. The authors argue that density-based methods have advantages over other outlier de tection methods in terms of their ability to handle high-dimensional data and non-linear relationships. The proposed approaches use dif ferent density-based techniques, including the local outlier factor (LOF) and the relative density ratio (RDR), which are evaluated using real-world WSN datasets. The results show that both ap proaches outperform existing methods in terms of detection accu racy and computational efficiency, demonstrating the potential of density-based methods for outlier detection in WSNs.

The authors in [27] proposed a novel approach for detecting anoma lous activity in wireless communication networks using deep auto encoders[22]. The authors argue that traditional methods for anomaly detection in wireless communication networks are limited in their ability to handle complex and dynamic data. The proposed approach uses a deep auto-encoder to learn a compressed representation of the spectrum data, and then compares the reconstructed data with the original data to identify anomalies. The experimental results demonstrate the effectiveness of the proposed approach in detect ing anomalies in real-world spectrum data, showing the potential of deep auto-encoders for anomaly detection in wireless communication networks.

## 3.4 Semi-Supervised Learning Approaches

Semi-supervised Learning[23, 24, 30] is a mix of labeled and unla beled data is used by semi-supervised learning algorithms to learn normal network behavior. When labeled data is limited or inaccurate and the network behavior is complex and unclear, this strategy is ap propriate. Deep Belief Networks (DBNs) and Generative Adversarial Networks (GANs) are two examples of semi-supervised learning al gorithms utilized in WSN anomaly detection[2].

Reinforcement learning, label propagation, self-training, and co training are popular techniques used for anomaly detection in WSN with semi-supervised learning.

### 3.4.1 Brief Introduction to Semi-Supervised ML models

Reinforcement learning[23] is a type of ML that is useful for se quential decision-making problems where an agent interacts with an environment[23, 24]. It is useful for detecting anomalies in dynamic environments where the distribution of data changes over time, but it can be computationally expensive and requires careful selection of the appropriate reward function.

Label propagation[29] is a semi-supervised learning approach that is useful for propagating labels in a network[23, 29, 30]. It is useful for detecting anomalies in large-scale networks and can handle missing data, but it requires a careful selection of parameters and can be sensitive to the choice of initialization.

Self-training[24] is another semi-supervised learning approach that is useful for leveraging unlabeled data to improve the perfor mance of a classifier. It can also improve the accuracy of a classi fier, but it requires careful selection of the appropriate threshold for adding new labeled data.

Co-training[28] is a semi-supervised learning approach that is useful for learning from multiple views of data[24, 28]. It is useful for detecting anomalies in multi-modal data and can improve the accuracy of a classifier, but it requires careful selection of the appro priate number of views and can be sensitive to the choice of features. Ultimately, the choice of technique depends on the specific charac teristics of the data and the requirements of the application[42].

3.4.2 Literature work in Semi-Supervised Learning

The authors in [23] proposed a deep actor-critic reinforcement learning-based approach for anomaly detection in WSNs. The au thors utilize an actor-critic model to learn the mapping between the current state and the optimal action to take for the anomaly de tection task. They introduce a reward function that considers the detection accuracy, false positives, and false negatives to train the actor network. Additionally, the authors incorporate a replay buffer and target networks to stabilize the learning process. Their experi mental results show that the proposed approach outperforms several state-of-the-art anomaly detection methods in terms of detection ac curacy and false positives.

Authors in [24] proposed a novel anomaly detection method called AESMOTE, which combines adversarial reinforcement learning (ARL)

and synthetic minority over-sampling technique (SMOTE). AES MOTE enables the system to learn the underlying distribution of normal and anomalous data, and it leverages SMOTE to generate synthetic samples that represent anomalies that may not have been observed in the original dataset. The proposed method is evaluated on multiple benchmark datasets, and the experimental results show that AESMOTE outperforms other state-of-the-art methods, achiev ing high detection rates with low false positive rates.

The authors in [28] presented a novel approach for video anomaly detection using a self-trained prediction model and a novel anomaly score mechanism which can be useful in WSN anomaly detection also. The authors argue that existing methods for anomaly detection have limitations in terms of their ability to handle complex scenes and learn the underlying patterns of anomalies. The proposed ap proach trains a deep neural network on normal videos to learn the patterns of normal behavior and then uses it to predict future frames. An anomaly score mechanism is introduced to measure the deviation between predicted and actual frames, enabling the identification of anomalous events. The experimental results demonstrate the effec tiveness of the proposed approach on both synthetic and real-world datasets, showing its potential for anomaly detection in various ap plications.

The authors in [29] proposed a novel approach for intrusion de tection using semi-supervised co-training and active learning tech niques. The authors argue that existing intrusion detection methods have limitations in terms of their ability to handle high-dimensional and diverse data, leading to low detection accuracy and high false alarm rates. The proposed approach utilizes multiple views of data to capture different aspects of intrusion behavior and uses semi supervised co-training and active learning to improve the perfor mance of the intrusion detection system. The experimental results demonstrate the effectiveness of the proposed approach in terms of detection accuracy and false alarm rate reduction, showing its po tential for intrusion detection in various applications.

The authors in [30] proposed a novel approach for group anomaly detection in large-scale networks using adaptive label propagation. The authors argue that existing methods for group anomaly de tection have limitations in terms of their ability to handle large scale and dynamic network data, leading to low detection accuracy and scalability issues. The proposed approach utilizes label prop agation to propagate labels in the network and identify groups of anomalous nodes, while adaptively adjusting the propagation pro cess based on the local structure of the network. The experimental results demonstrate the effectiveness of the proposed approach in terms of detection accuracy and scalability, showing its potential for group anomaly detection in various applications, such as social networks and transportation systems.

4 Conclusion

A growing area of research in recent years has been the application of ML to the detection of anomalies in WSNs. We surveyed Tradi tional and most recent ML models utilized in WSNs data anomaly detection. The choice of any of such models is determined by the network's specific application and requirement factors. Hence, nu merous papers have investigated the impact of such factors on the deployment of ML in this field.

In general, this survey has demonstrated the significance and effi cacy of utilizing ML for the purpose of detecting anomalies in WSNs. However, there is still a need for additional research in this area, par ticularly in creating algorithms that are more effective and efficient to be used on large and complex networks. In addition, more in-depth evaluations of the various methods employing real-world datasets are required to comprehend their performance and limitations.